\documentclass[10pt,twocolumn,letterpaper]{article}
\usepackage[table]{xcolor}

\usepackage{cvpr}              

\definecolor{cvprblue}{rgb}{0.21,0.49,0.74}
\usepackage[pagebackref,breaklinks,colorlinks,citecolor=cvprblue]{hyperref}
\usepackage{multirow}
\def\paperID{*****} 
\def\confName{CVPR}
\def\confYear{2024}

\usepackage{hyperref}
\title{Multi-Modal UAV Detection, Classification and Tracking Algorithm---Technical Report for CVPR 2024 UG2 Challenge}

\author{Tianchen Deng\textsuperscript{1}, Yi Zhou\textsuperscript{2}, Wenhua Wu\textsuperscript{1}, Mingrui Li\textsuperscript{3}, Jingwei Huang\textsuperscript{4}, Shuhong Liu\textsuperscript{5}, Yanzeng Song,\\ Hao Zuo\textsuperscript{3},  Yanbo Wang\textsuperscript{1},Yutao Yue\textsuperscript{2,6}, Hesheng Wang\textsuperscript{1},Weidong Chen\textsuperscript{1}
\\
{\textsuperscript{\rm 1} Shanghai Jiao Tong University}
{\textsuperscript{\rm 2} Institute of Deep Perception Technology, JITRI} \\
{\textsuperscript{\rm 3} Dalian University of Technology}
{\textsuperscript{\rm 4} University of Electronic Science and Technology of China}\\
{\textsuperscript{\rm 5} The University of Tokyo}
{\textsuperscript{\rm 6} The Hong Kong University of Science and Technology (Guangzhou)}
}

\begin{document}
\maketitle
\begin{abstract}
This technical report presents the 1st winning model for UG2+, a task in CVPR 2024 UAV Tracking and Pose-Estimation Challenge. This challenge faces difficulties in drone detection, 
 UAV-type classification, and 2D/3D trajectory estimation in extreme weather conditions with multi-modal sensor information, including stereo vision, various Lidars, Radars, and audio arrays. Leveraging this information, we propose a multi-modal UAV detection, classification, and 3D tracking method for accurate UAV classification and tracking. A novel classification pipeline which incorporates sequence fusion, region of interest (ROI) cropping, and keyframe selection is proposed. Our system integrates cutting-edge classification techniques and sophisticated post-processing steps to boost accuracy and robustness. The designed pose estimation pipeline incorporates three modules: dynamic points analysis, a multi-object tracker, and trajectory completion techniques. Extensive experiments have validated the effectiveness and precision of our approach. In addition, we also propose a novel dataset pre-processing method and conduct a comprehensive ablation study for our design. We finally achieved the best performance in the MMUAD dataset in classification and tracking. The code and configuration of our method are available at  \href{https://github.com/dtc111111/Multi-Modal-UAV}{https://github.com/dtc111111/Multi-Modal-UAV} .
\end{abstract}    
\section{Introduction}
Unmanned aerial vehicles (UAVs), commonly referred to as drones, have become increasingly accessible and played an important role in various fields, such as transportation~\cite{plgslam,prosgnerf,liu3}, photography~\cite{xie,xie2,zhu1,xing1,xing2,li3}, and search~\cite{wang1,weng1}, bringing considerable benefits to the general public. However, the proliferation and capabilities of small commercial UAVs have also introduced multifaceted security challenges that extend beyond conventional concerns.

In recent years, there has been a significant surge in research focused on anti-UAV systems. Despite this interest, the majority of existing systems operate on a single-modal basis. The UG2+ Challenge at CVPR 2024 aims to push the boundaries by requiring participants to develop a novel multi-modal anti-UAV system. This challenge involves the joint estimation of UAV 3D trajectories and the identification of UAV types. 
The UG2+ competition organizers collect their own dataset: MMAUD dataset~\cite{mmaud}. The core of this challenge is how to effectively utilize multi-modal information to achieve both robust 3D UAV position estimation and UAV type classification, even in challenging conditions where single sensors may fail to acquire valid information.
For the classification task, the key challenge is when drones operate at high altitudes or encounter extreme visual conditions. Existing methods struggle to detect small drones due to their compact size, resulting in a smaller visual presence. For the tracking part, it is also difficult to detect and estimate the 3D position of the small drone with the reduced Radar Cross Section, noisy lidar points, and interference from surrounding dynamic objects.

 To this end, we propose our multi-modal method, which effectively leverages various Lidars and camera information. There are two parts in our network: the classification network and the pose estimation pipeline. For the classification network, We first carry out pre-processing and serialization of the dataset to implement data augmentation, in order to cope with adverse weather conditions.  The 3D UAV pose estimation method proposed leverages lidar data due to the unreliability of visual depth information and radar data. Instead of direct network training, a pipeline is devised to exploit features such as spatial density, motion signature, and trajectory smoothness in an unsupervised manner, supplementing label-provided semantic information. \textbf{Overall, our contributions are shown as follows:}
\begin{itemize}
    \item 
    We propose the first multi-modal UAV classification and 3D pose estimation method for accurate and robust anti-UAV system.
    \item  
    A novel classification pipeline is introduced, incorporating sequence fusion, region of interest (ROI) cropping, and keyframe selection. Our system integrates advanced classification techniques and post-processing steps to enhance accuracy and robustness. Extensive experiments validate the effectiveness and precision of our approach.
    \item A pose estimation pipeline is proposed with dynamic points analysis, multi-object tracker, and trajectory completion. Extensive experiments demonstrate the effectiveness and accuracy of our system. We achieve the best performance in UG2+ challenge in CVPR 2024. 

\end{itemize}

\begin{figure*}
    \centering\includegraphics[width=\linewidth]{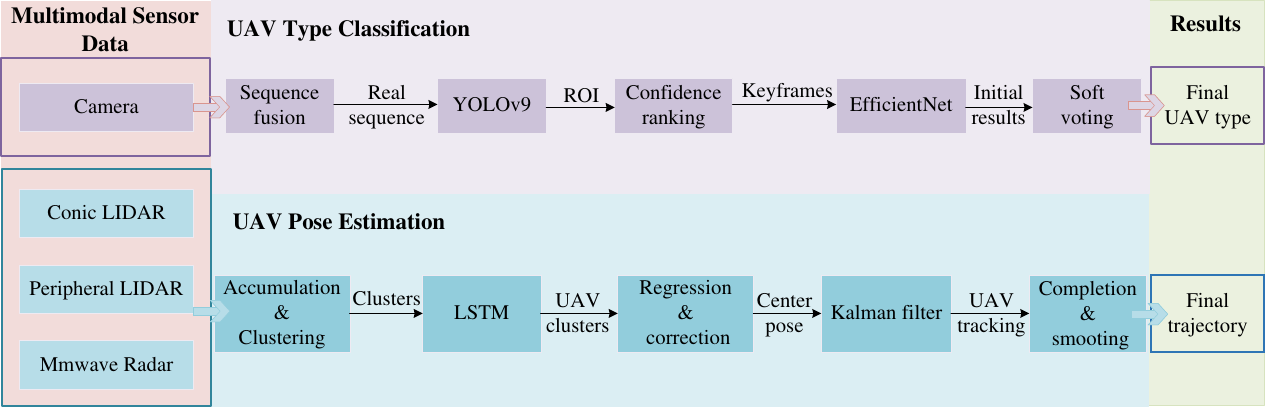}
    \caption{Overview of multi-modal UAV classification and tracking framework.}
    \label{fig:framework}
\end{figure*}

\section{Related Work}


\noindent \textbf{Detection and Classification.}
UAV detection~\cite{qiao1,qiao3,panfeng2} and classification techniques~\cite{tw2,shen1,zhu2} have seen significant advancements with the integration of deep learning methodologies across various sensor modalities. High resolution range sensor, such as MIMO radar~\cite{klare2017uav} and Lidars~\cite{hammer2018lidar}, which directly utilize the point cloud for classification. In addition, radar-based detection systems utilize the micro-Doppler effect~\cite{de2012micro} to identify UAVs by their unique rotational patterns. Vision-based detection systems utilize neural networks to process visual data from cameras. Models from the YOLO series~\cite{yolov7,yolov8,yolov9} exhibit high accuracy in bounding box classification and regression. Liu.~\cite{liunote} propose a enhance detection and classification methdos using clustering SVM, achieving better performance. Some methods use segmentation methods\cite{liucite,panfeng1,panfeng4,tw1,qiao2} to improve the detection performance.

Despite these advancements, challenges persist due to the diversity in UAV sizes and dynamic behaviors. Environmental factors like weather conditions, background noise, and the presence of other wireless signals can interfere with detection systems~\cite{zhou2022towards}. Addressing these challenges requires a sensor fusion framework. Some methods estimate the vehicle pose and shape~\cite{zhang1}. Some methods harness the capabilities of multiple sensor types, blending their strengths to enhance the robustness and accuracy of UAV detection and classification. For example, integrating radar and vision data combines the radar's ability to penetrate adverse weather conditions with the high-resolution imaging provided by cameras. This multi-sensor fusion approach has been successfully applied in various research efforts, such as those documented in studies like ~\cite{zhao2023lif,xin2,xin3,xin1}, demonstrating superior performance compared to systems relying on a single sensor.

\noindent \textbf{3D Tracking.}
 UAV 3D tracking~\cite{panfeng3} has various application in real-world such as military, transportation~\cite{lh1,lh2,lh3}, and security~\cite{weng1,weng2}. Some systems integrate the Bayesian tracking framework, employing techniques like Kalman filter ~\cite{kf1} and particle filter~\cite{pf1} to maintain robust tracking performance.  Some methods use learning-based method to improve the accuracy. Lan et al.~\cite{tracking1} applied the sparse learning method to RGB-T tracking, thereby removing the cross-modality discrepancy. Liu et al.~\cite{tracking2} proposed a mean-shift-based method which transformed the target position to 3D coordinates using RGB and depth images. Moreover, the development of advanced algorithms for data fusion, such as the use of deep neural networks for feature extraction and decision-level fusion, has been a significant area of progress. These algorithms can learn complex relationships and correlations between data from different sensors, resulting in a more comprehensive understanding of the environment~\cite{qin2023motiontrack, qin2023supfusion}. Some SLAM methods~\cite{neslam,ram,li1,li2,compact} are also used in 3D tracking method to further improve the accuracy.

Additionally, the transformer-based algorithms~\cite{liu1,liu2} for multi-object tracking could be adapted for UAV detection scenarios. These algorithms~\cite{vaswani2017attention}, originally from the field of natural language processing, have shown their effectiveness in handling complex data associations and could potentially improve the tracking of UAVs in multi-sensor environments~\cite{zhang2023motiontrack}. Some methods use joint learning methods~\cite{incremental,zhu11,wyh1,wyh2} to learn the UAV pose and the type simultaneously.

\section{Method}

Our algorithm workflow is illustrated in Fig.~\ref{fig:framework}. The input multi-modal sensor data is processed through two pipelines: UAV type classification and 3D tracking. The UAV type classification pipeline primarily utilizes image data, while the UAV pose estimation pipeline primarily utilizes data from Lidars and radar. In this section, we will describe the details.

\subsection{UAV Type Classification}

The trajectory of UAV movement typically exhibits continuity, facilitating the utilization of contextual information for training the classifier. Additionally, upon examination of the dataset, we identified numerous sequences as subsets of a broader sequence, termed as \textit{real sequences}. Furthermore, our analysis revealed that sensor data concerning UAVs often exhibits sparsity, with merely a few pixels or points belonging to the targets at high altitudes. This inherent constraint significantly increases the difficulty of single-frame classification.

Previous sequential tasks usually extend the model through the temporal axis or employ a transformer architecture. However, for the MMAUD dataset, most frames do not provide valid information as mentioned above. Therefore, we adopt the following strategy to accomplish the classification task in three steps: sequence fusion grounded in feature similarity, region of interest cropping, and keyframe selection utilizing YOLOv9-e, classification, and post-processing.

\subsubsection{Sequence Fusion Based on Feature Similarity}
We have formulated two assumptions to guide the construction of the \textit{real sequences}. Firstly, each \textit{real sequence} contains only one UAV. Secondly, there exists a substantial temporal gap between consecutive \textit{real sequences}.

By observing the foreground and background, we sample the data from the original sequence at a ratio of 1/100. Subsequently, we leverage the EfficientNet-B7 \cite{effnet} pre-trained on ImageNet to extract feature representations from the sampled images. Feature aggregation is accomplished by averaging the representations extracted from the sampled images. Then we extract representational features from each original sequence, compute the cosine similarity of each representation, and apply a threshold to group sequences to construct the \textit{real sequences} in both the training and test phases.

\subsubsection{ROI Crop and Keyframe Selection Based on YOLOv9}
During the training phase, we apply YOLOv9-e \cite{yolov9} without finetuning on all images of the \textit{real sequences} with selection of airplane to get the ROI of UAV. Although there are misclassifcations in zero-shot results, this process still helps us automatically select enough UAV images to train the classifier.  Following the cropping and rescaling of the ROI from the detection results, we additionally utilize a random sampling procedure to mitigate class imbalance during training ($N \leq 300$ for each \textit{real sequence}).

During the test phase, we also employ YOLOv9-e \cite{yolov9} to detect the UAV in a zero-shot manner, and rank the confidence scores for detection == ``airplane''. Ideally, we aim to use the most clearly detected image for per-image classification as the prediction for the entire \textit{real sequence}, assuming each sequence contains the same type of UAV. While confidence does not directly indicate how clear the UAV image is, it does reflect the model's confidence. Therefore, we use this metric to identify $k$ keyframes and employ a soft classification strategy by aggregating the softmax probabilities from these keyframes.

\subsubsection{Classification and Post-Processing}
The training dataset is small because only UAVs close to the ground ($\le 10 m$) and centered in the camera's view can be effectively detected. Therefore, the detection networks are supposed to be light-weight models. Here, we train EfficientNet-B7 \cite{effnet} to obtain the initial results.

During the test phase, we apply our classification model on each keyframe of the \textit{real sequence} and add the softmax results of each sequence to form a soft majority vote strategy. Finally, we retrieve the data sequence predictions from the \textit{real sequences}.


\begin{figure*}
    \centering
    \includegraphics[width=\linewidth]{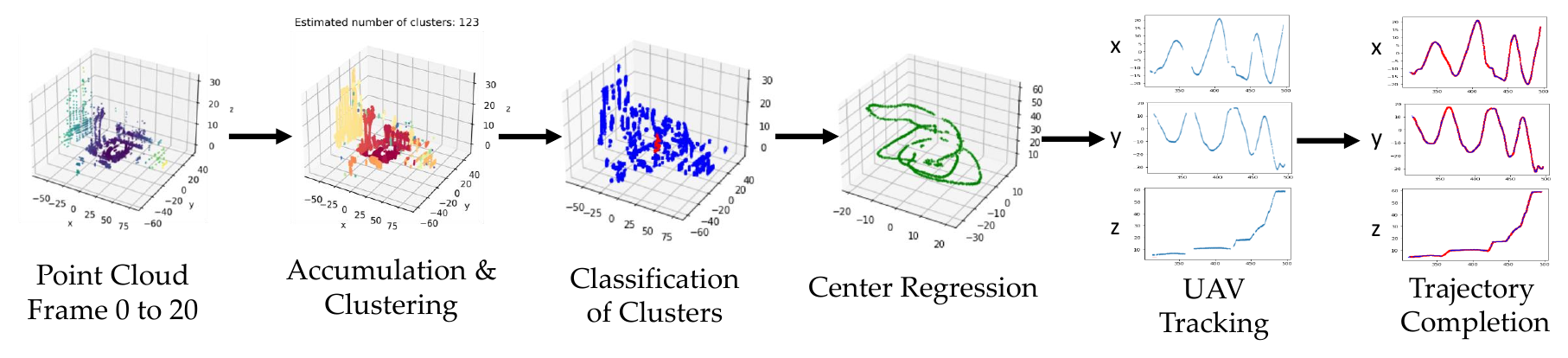}
    \caption{The UAV pose estimation pipeline. Initially, we accumulate and cluster data over 20 frames, then proceed with an analysis of the dynamic point cloud to segregate background points from dynamic points. Subsequently, we track these dynamic points and perform center regression on dynamic clusters to derive an initial estimation of the UAV trajectory. The trajectory is further refined and extrapolated using a Kalman filter, culminating in a comprehensive and precise UAV trajectory.}
    \label{fig:tracking}
\end{figure*}

\begin{figure}
    \centering
    \includegraphics[width=\linewidth]{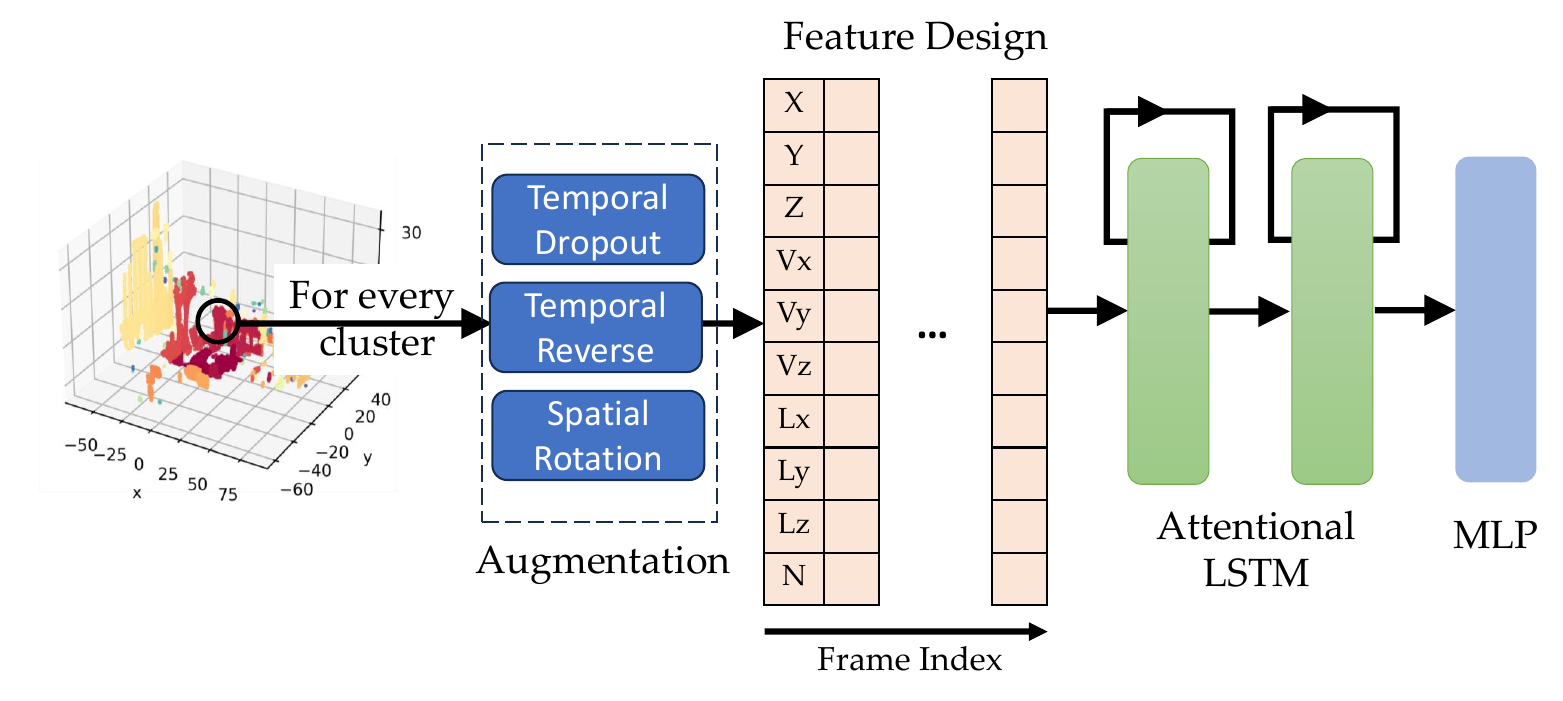}
    \caption{ The pipeline of dynamic point analysis module. We first accumulate 20 frames of point clouds as a temporal window. Then we use an unsupervised clustering method to cluster the point clouds and extract the feature of each group. The LSTM module with MLP head can finally decompose the dynamic points.}
    \label{fig:tracking2}
\end{figure}
\subsection{UAV Pose Estimation}
Tracking UAVs in complex weather conditions and at high altitudes presents significant challenges. Given the unreliability of visual depth information at long distances, our pose estimation method primarily utilizes point cloud data from Lidar and radar. Upon examining the dataset, it became apparent that although UAV flight trajectories exhibit high diversity, the acquisition environment remains fairly uniform, and the labeling only includes point annotations for the UAVs. Consequently, rather than directly training a neural network for pose estimation, we suggest a pipeline that investigates features such as spatial density, motion signatures, temporal consistency, and trajectory smoothness in an unsupervised manner to enrich the semantic details provided by the labels. Our pose estimation pipeline comprises three key modules: dynamic point cloud analysis, a multiple object tracking module, and trajectory completion. The pipeline of our pose estimation framework is shown in Fig.~\ref{fig:tracking}.

\subsubsection{Dynamic Point Analysis}

When UAVs operate at higher altitudes, they constitute only a small portion of the overall point cloud. Directly training a segmentation network on this data tends to yield suboptimal outcomes, such as classifying all points as background. Noting that UAVs generally operate in clear skies away from other objects, we design a two-stage approach to address these challenges. First, we employed an unsupervised clustering method to cluster the point cloud data. Then, we extract relevant geometric and motion features from these clusters to decompose the dynamic points. The framework of our dynamic point analysis method is illustrated in  Fig.~\ref{fig:tracking2}. 

We begin by accumulating 20 frames of point clouds to create a temporal window. Within this window, we extract motion features, such as the velocity vector of the center points. We incorporate temperal dropout, temperal reverse, and spatial rotation to augment the point cloud. Then, we extract seven-dimensional features. Subsequently, we design a network that includes an attention-based LSTM module for temporal analysis, a Multi-Layer Perceptron (MLP) for sequence classification, and a PointNet-based module for center regression of the detected UAV clusters. This comprehensive approach allows for precise tracking and classification of UAVs.

Instead of relying on the last hidden state of the LSTM module alone to encapsulate the sequence, we have integrated an attention mechanism that combines all hidden states into a comprehensive representation. This mechanism dynamically assigns importance weights to each of the hidden states, effectively combining them into a single weighted feature set. By selectively highlighting the most salient elements, the model's ability to cope with long sequences and recognise complex motion patterns is greatly enhanced. 

Cluster classification is conducted using an MLP head, with ground truth generated through the nearest neighbor association between the UAV pose labels and the estimated cluster centers. To mitigate overfitting, we have implemented a range of point cloud augmentation techniques. These include global rotation, temporal reversion, and frame dropout, which enhance the model's robustness through spatial-temporal augmentation.


While the cluster center can initially be predicted as the geometric mean within the cluster, issues such as incomplete point clouds and potential distance-related measurement biases in the dataset necessitate a more robust approach. Therefore, we employ an additional MLP specifically for the task of center regression. Our observations indicate a strong correlation between the regression error and the cluster center in the training data. To tackle this, we develop a nonlinear model for bias correction. Specifically, we use a third-order polynomial feature transformer to expand the three-dimensional coordinates into a 24-dimensional feature space. We then perform linear regression to delineate the relationship between these expanded features and the observed bias. The corrected cluster center is determined by adjusting the initial estimate with the predicted bias, enhancing the accuracy of our localization.


\subsubsection{Multiple Object Tracking}

In the detection process, there exists an inherent trade-off between accuracy and recall, often leading to results that include both clutter and missed detections. Additionally, despite corrections, the predicted cluster center may exhibit a zig-zag pattern, particularly in sparse point clouds. To address these challenges, we implement a multiple object tracker that helps filter out clutter and smoothens trajectories. We employ the linear Kalman filter as the backbone of our tracking framework. Within this framework, new tracks are initiated from unassociated measurements, and existing tracks are terminated when their covariance exceeds a predefined threshold. This approach enhances the clarity and reliability of the tracking outcomes.

Since the effectiveness of our proposed center regression module, we set low diagonal values in the noise covariance matrix. In addition, since we prioritize recall over accuracy in our classification module, there could be some clutters which are wrongly classified as UAV. Therefore, we set the association threshold low so that the predicted trajectories are robust to the possible clutter from the detector. 

\subsubsection{3D Trajectory Completion and Smoothing}

Given the strict threshold applied in the deletion process, the estimated trajectories might appear fragmented. However, for pose estimation tasks, we can access entire trajectories without causal constraints. This access to contextual information allows for more effective trajectory prediction and completion, which helps address issues related to missed detections and lost trackers. To enhance the trajectory continuity, we employ a third-order autoregressive (AR) model for trajectory completion. This model utilizes data from the previous three time steps to predict the subsequent step in the sequence, thereby providing a more cohesive and continuous trajectory estimation.

When faced with missing observations or lost trackers in the input data, the autoregressive (AR) model is capable of generating predictions using the available information. However, the absence of data introduces uncertainty and can impact the accuracy of these predictions. To mitigate these effects, we interpolate the predicted trajectories according to the specific timestamps of the test data and apply smoothing techniques to enhance the trajectory's continuity and accuracy.

Considering that UAVs commonly employ spline approximation in their path planning modules, we choose B-spline interpolation for the smoothing process. This method is particularly well-suited for creating smooth and flexible trajectories, making it ideal for adapting the UAV's flight path to the dynamic conditions typically encountered during operations. This approach helps ensure that the trajectory remains reliable and precise, even in the presence of data gaps.

\section{Experiment}

\subsection{MMUAD Dataset Analysis}

The competition is based on a subset of data from the comprehensive MMAUD dataset. Participants are tasked with classifying four types of drones: Phantom 4, M300, M30T, and Mavic 3, and estimating their pose. The dataset provides unsynchronized measurements from various sensors, including stereo fisheye cameras, two types of Lidars (conic 3D Lidar and peripheral 3D Lidar), and 4D millimeter-wave radar. For training, the competition offers 102 training sequences and 16 validation sequences, each lasting approximately 20 seconds and 5 seconds, respectively. The final evaluation is conducted on a test set consisting of 59 sequences. The ranking in this challenge is determined based on two criteria: i) Mean Square Error (MSE Loss) compared to the ground truth labels of the test set, and ii) the classification accuracy of UAV types in the test set.

The dataset employs a sensor rig consists of four sensors, including
\begin{itemize}
    \item  Stereo Fisheye Cameras: These cost-effective cameras offer a panoramic 180-degree field of view (FoV), creating a dome-shaped detection volume that is instrumental in horizon scanning and providing a wide coverage area for UAV detection. Their affordability and wide FoV make them ideal for continuous surveillance.
    \item Conic 3D Lidar: This upward-facing Lidar has a 70-degree conic FoV and is adept at detecting objects at distances of up to 300 meters. Its conic scanning pattern complements the fisheye cameras by focusing on a central area and extending the range of detection beyond the visual capabilities, ensuring that distant drones are captured.
    \item    Peripheral 3D Lidar: With a 360-degree horizontal and a 59-degree vertical FoV, this Lidar provides comprehensive peripheral coverage on the ground level, effectively detecting nearby threats within a 70-meter range. It collaborates with the conic Lidar to ensure that the detection system has no blind spots and covers both close and far distances.
    \item Mmwave Radar: Operating at 77GHz, this radar boasts a 120-degree horizontal and a 30-degree vertical FoV, capable of sensing moving objects at distances of up to 350 meters. The radar's ability to detect motion is particularly valuable, as it can track the trajectory of drones and is less affected by environmental conditions like lighting or weather.

\end{itemize}

Together, the combination of these sensors expands the perception field-of-view. The stereo fisheye cameras provide broad situational awareness, while the conic and peripheral Lidars offer detailed detection at varying distances. The mmWave radar enhances the system's ability to track moving targets over a considerable range. The primary challenge of this dataset lies in effectively harnessing the complementary information from the four types of sensors to achieve robust perception.

\begin{figure}
    \centering
    \includegraphics[width=\linewidth]{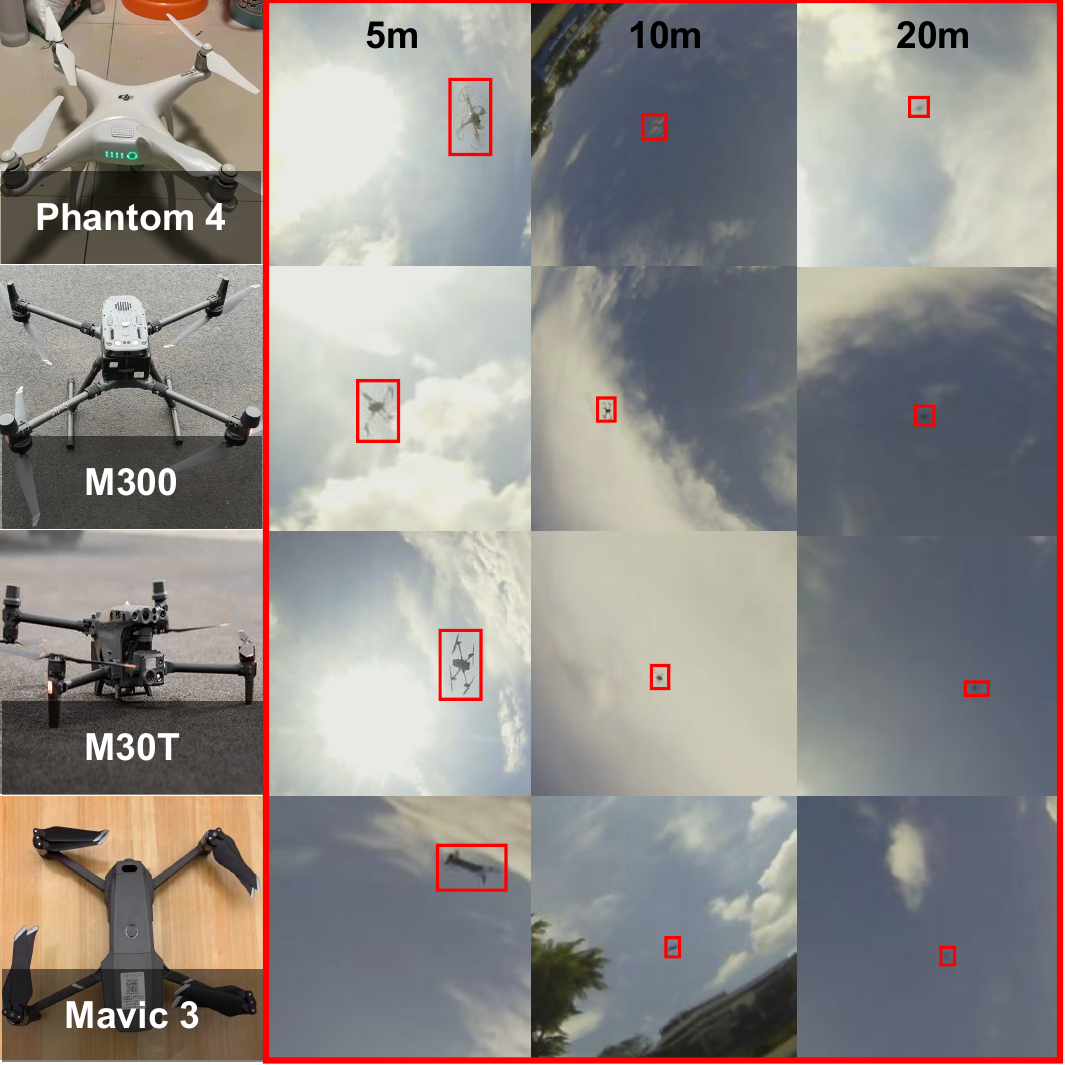}
    \caption{
We visualize the distribution of images in the test datasets. }
    \label{fig:dataset}
\end{figure}

\begin{figure}
    \centering
\includegraphics[width=\linewidth]{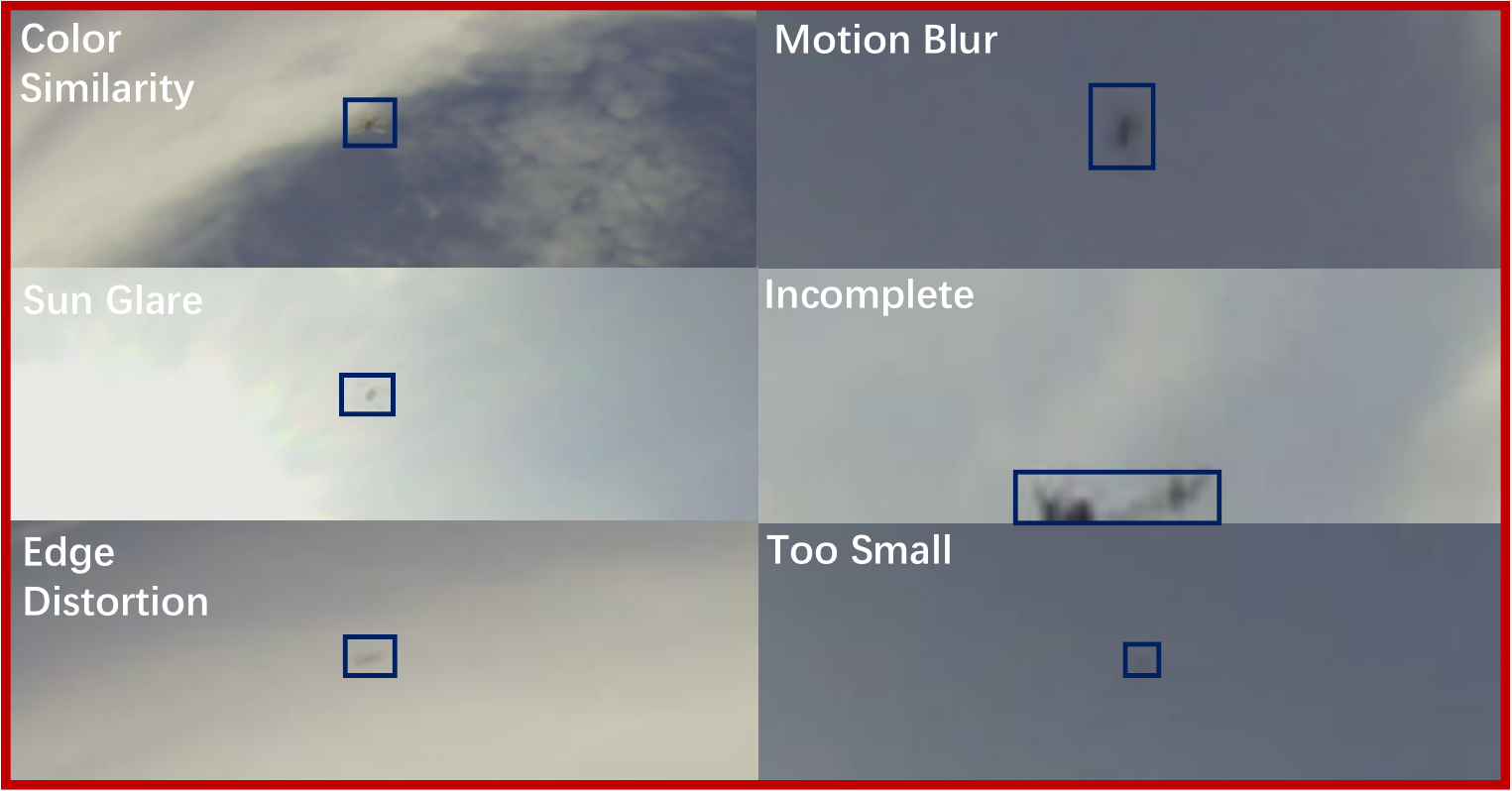}
    \caption{Some corner case in images.}
    \label{fig:cornercase}
\end{figure}

In Fig.~\ref{fig:dataset}, we visualize some example images of four types of drones at three different altitudes (5m, 10m, 20m).  At low altitudes, the UAVs are clearly identified at 5 meters with clear details for classification. However, at altitudes over 20 meters, the UAV becomes a point target or even unobservable, thus making visual classification difficult. Fig.~\ref{fig:cornercase} highlights some challenging cases for low-altitude visual detection, including color similarity, motion blur, sun glare, small objects, incomplete objects, and edge distortion of fish-eye camera.

\begin{table*}[htbp]
  \centering
  
  \label{tab:example}
  \begin{tabular}{c|cccc}
    \hline
    Rank & Participant & Entries & Pose MSE Loss $\downarrow$ & UAV Type classification Accuracy $\uparrow$ \\
    \hline
    \rowcolor{orange!80} 1&Ours       &19 &2.21375   & 0.8136\\
    \rowcolor{orange!50} 2&Gaofen\_Lab &7  &7.299575 & 0.322\\
    \rowcolor{orange!20} 3&sysutlt    &39 &24.50694 & 0.322 \\
    4&casetrous  &1  &56.880267 & 0.2542 \\
    5&NTU-ICG    &7  &120.215107 & 0.322\\
    6&MTC        &26 &189.669428 & 0.2724\\
    7&gzist      &1  &417396317 & 0.2302\\
    \hline
  \end{tabular}
  \caption{Competition ranking list in test dataset. We achieve 
 the best performance in both pose estimation accuracy and UAV type classification accuracy, and we are far ahead of other groups.}
 \label{tab:ranking}
 \vspace{-0.4cm}
\end{table*} 


\begin{figure}
    \centering\includegraphics[width=\linewidth]{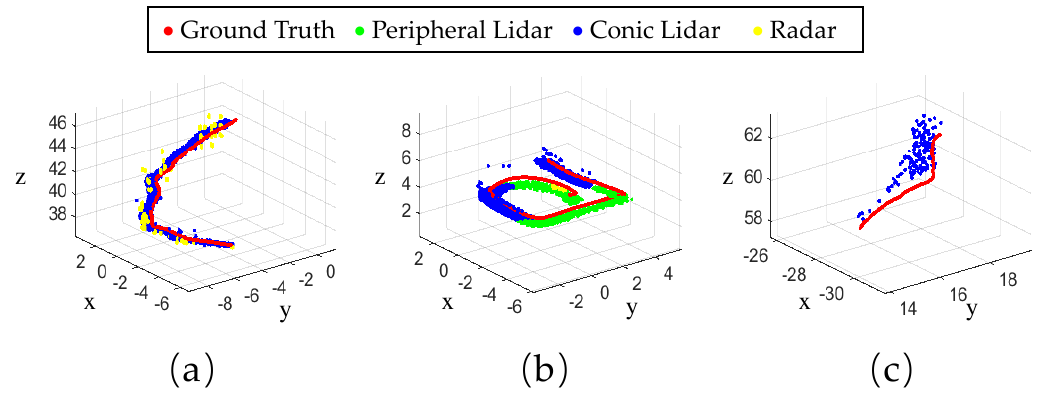}
    \caption{Some examples of point cloud measurements: (a) is a typical case where both radar and conic Lidar return measurements. (b) is the cases where the UAV crosses through the FoVs of two Lidars (c) shows the bias in the measurement in the high altitude. }
    \label{fig:cornercase2}
    \vspace{-0.4cm}
\end{figure}

In Fig.\ref{fig:cornercase2}, we visualize example sequences of point cloud measurements. In Fig.\ref{fig:cornercase2} (a), we observe the co-existence of measurements from both conic Lidar and radar. However, it is evident that the conic Lidar produces a higher quality point cloud compared to the radar, which appears sparse and inconsistent. In Fig.\ref{fig:cornercase2} (b), we notice the UAV crossing the field of view of two Lidars. In Fig.\ref{fig:cornercase2} (c), we observe position bias at high altitudes due to insufficient resolution. These challenges inspire the design of our pose estimation pipeline as late fusion.

\subsection{Experimental Results}
We present the 3D pose estimation performance and UAV type classification performance of our system on test dataset. In Tab.~\ref{tab:ranking}, we can see that we achieve the best performance on 3D tracking and UAV type classification. Our UAV classification method successfully fuses information across sequences, utilizing a soft vote strategy to accurately identify the type. 

\begin{figure}
    \centering
    \includegraphics[width=\linewidth]{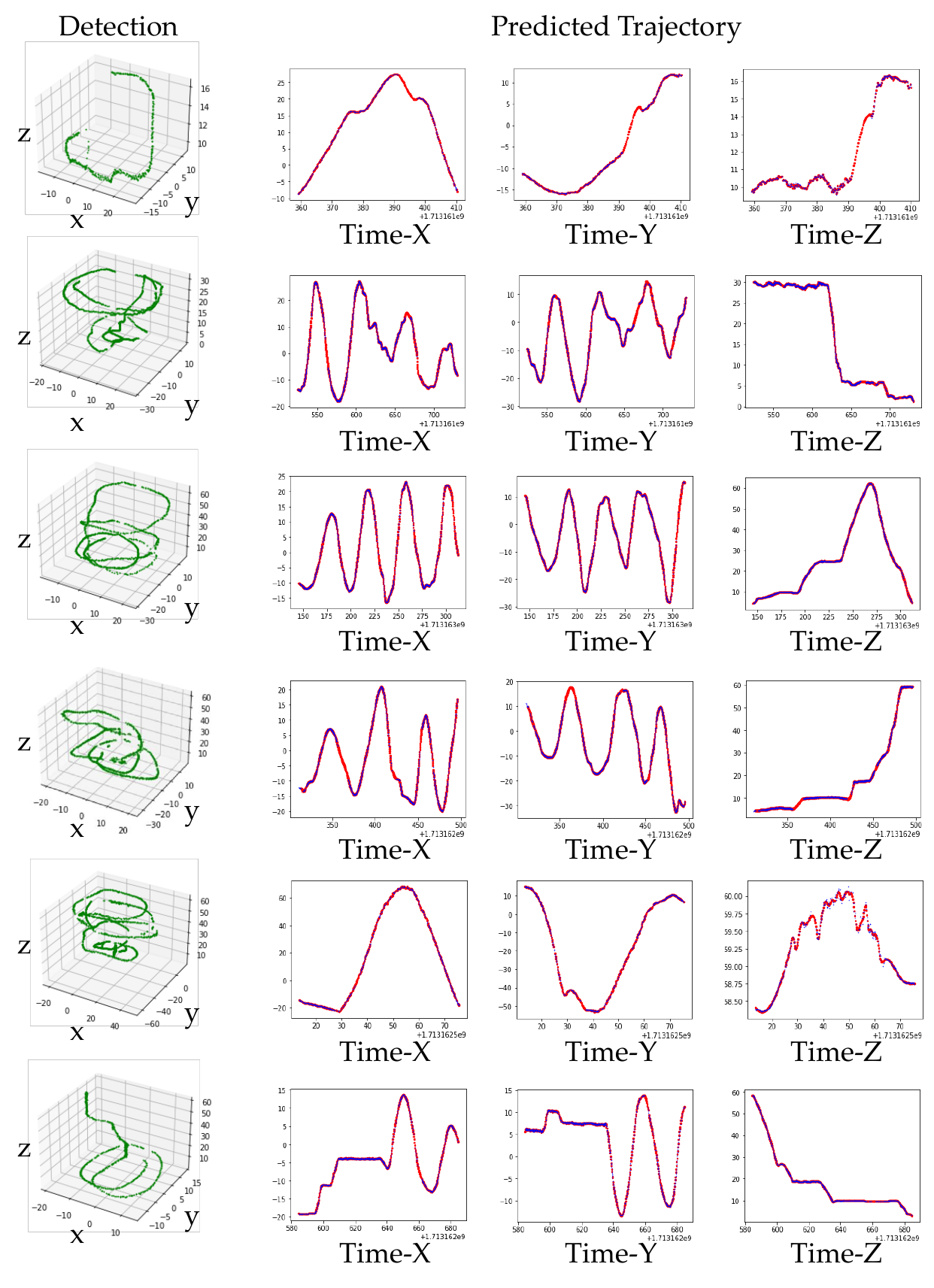}
    \caption{Detection results and predicted trajectories for the test set: The red curves are the trajectories given by the tracker and the blue curves are the trajectories after completion.}
    \label{fig:pose_results}
    \vspace{-0.4cm}
\end{figure}

For the point cloud-based UAV detector, the validation accuracy is 0.9998 and the recall is 0.9184. For the center regression task, the MSE loss decreased from 0.27 to 0.05. These results indicate that our lightweight detection framework can successfully detect UAVs and predict the cluster centers. The detection results for the test sequences are shown in \cref{fig:pose_results}. We can see that there are some noisy detections and missed trajectories. After applying the multiple object tracker, the noisy detections are filtered out and the trajectory is smoothed, as shown by the red curves. Finally, the missed trajectories are interpolated using the contextual information, as shown by the blue trajectories. As shown in Tab.~\ref{tab:ranking}, the pose MSE is 2.21 on the test dataset. The performance gap between the validation and test sets is due to extrapolation error, where the trajectories in the initial or ending stages are missed by our detector for a long period. In such cases, the contextual information is insufficient to infer the trajectory, leading to accumulated errors and thus a larger error. In future work, we will incorporate the data-driven prediction module into the entire framework to address this limitation.

\section{Conclusion}
In summary, we propose the first multi-modal anti-UAV system, achieving accurate 3D UAV tracking and UAV type classification. The multi-modal dataset pre-processing and sequential method significantly improve the classification performance. The proposed tracking module with dynamic point analysis, multi-head tracking, and 3D trajectory prediction further improve the UAV tracking accuracy. As a result, we finally achieve the 1st place in Ug2+ challenge in CVPR 2024. We hope our system can provide new insights and ideas to professionals involved with multi-modal anti-UAV systems.
{
    \small
    \bibliographystyle{ieeenat_fullname}
    \bibliography{main}
}


\end{document}